\documentclass[11pt,a4paper]{article}
\usepackage[hyperref]{emnlp-ijcnlp-2019}
\usepackage{times}
\usepackage{latexsym}
\usepackage{amsmath}

\usepackage{url}

\aclfinalcopy

\title{Biomedical relation extraction with pre-trained language representations and minimal task-specific architecture}

\author{Ashok Thillaisundaram \\
  BenevolentAI \\
  4-6 Maple Street\\
  Bloomsbury, London\\
  W1T 5HD\\
  {\tt ashok@benevolent.ai} \\\And
  Theodosia Togia \\
    BenevolentAI \\
  4-6 Maple Street\\
  Bloomsbury, London\\
  W1T 5HD\\
  {\tt sia@benevolent.ai} \\}

\date{}

\begin{document}
\maketitle
\begin{abstract}

This paper presents our participation in the AGAC Track from the 2019 BioNLP Open Shared Tasks. We provide a solution for Task 3, which aims to extract ``gene -- function change -- disease" triples, where ``gene" and ``disease" are mentions of particular genes and diseases respectively and ``function change" is one of four pre-defined relationship types. Our system extends BERT \cite{bert}, a state-of-the-art language model, which learns contextual language representations from a large unlabelled corpus and whose parameters can be fine-tuned to solve specific tasks with minimal additional architecture. We encode the pair of mentions and their textual context as two consecutive sequences in BERT, separated by a special symbol. We then use a single linear layer to classify their relationship into five classes (four pre-defined, as well as `no relation'). Despite considerable class imbalance, our system significantly outperforms a random baseline while relying on an extremely simple setup with no specially engineered features.
\end{abstract}

\section{Introduction}

Bidirectional Encoder Representations from Transformers (BERT) is a language representation model that has recently advanced the state of the art in a wide range of NLP tasks (e.g. natural language inference, question answering, sentence classification etc.) \cite{bert}. This is due to its capacity for learning lexical and syntactic aspects of language \cite{what-does-bert-look-at,bert-syntactic-abilities} using large unlabelled corpora. BERT achieves much of its expressive power  using a bi-directional Transformer encoder \cite{attention-is-all-you-need} and a `predict the missing word" training objective based on  Cloze tasks \cite{cloze-procedure}. In the biomedical domain, BioBERT \cite{biobert} and SciBERT \cite{scibert} learn more domain-specific language representations. The former uses the pre-trained BERT-Base model and further trains it with biomedical text (Pubmed\footnote{\url{https://www.ncbi.nlm.nih.gov/pubmed/}} abstracts and Pubmed Central\footnote{\url{https://www.ncbi.nlm.nih.gov/pmc/}} full-text articles). The latter trains a BERT model from scratch on a large corpus of scientific text (over 80\% biomedical) and learns a domain-specific vocabulary using WordPiece tokenisation \cite{wordpiece}. 

BERT has been adapted for use in relation extraction as a basis for supervised, unsupervised and few-shot learning models \cite{matching-the-blanks}. A recent model, Transformer for Relation Extraction (TRE) \cite{tre-transformer-re} uses an architecture similar to that of BERT by extending the OpenAI Generative Pre-trained Transformer \cite{radford2018improving}, in order to perform relation classification for entity mention pairs. In contrast to BERT, TRE uses a next word prediction objective. The model encodes the pairs and their context in a sequence separated by a special symbol. In our model, we use a similar way of encoding gene-disease pairs and their textual context in order to predict their `function change' relationship, but in contrast to TRE, we leverage SciBERT's domain-specific vocabulary and representations learnt from scientific text.

\section{Task and data}

\paragraph{Task description}

Task 3 of the AGAC track of BioNLP-OST 2019 involves Pubmed abstract-level relation extraction of gene-disease relations. The relations of interest concern the function change of the gene which affects the disease. The four relation types are:
\begin{itemize}
    \item Loss of Function (LOF): a gene undergoes a mutation leading to a loss of function which then affects the disease.
    \item Gain of Function (GOF): a gene mutation causes a gain of function.
    \item Regulation (REG): the function change is either neutral or unknown.
    \item Complex (COM): the function change is too complex to be described as any of the former relations.
\end{itemize} 

To illustrate these relation types more concretely, we repeat the examples given on the task webpage. The following sentence depicts the Gain of Function relation between SHP-2 and juvenile myelomonocytic leukemia: \textit{`Mutations in SHP-2 phosphatase that cause hyperactivation of its catalytic activity have been identified in human leukemias, particularly juvenile myelomonocytic leukemia.'}
In this case, `hyperactivation of catalytic activity' indicates Gain of Function. 

An example of the Regulation relation, on the other hand, would be the following sentence: \textit{`Lynch syndrome (LS) caused by mutations in DNA mismatch repair genes MLH1.'}. The phrase `caused by' demonstrates an association between MLH1 and Lynch syndrome but no information is given on the specific nature of the mechanism relating them.

\paragraph{Annotated corpus}

The training data provided consist of 250 PubMed abstracts with annotations of the form `Gene; Function change; Disease' for each abstract. For test data, a further 1000 PubMed abstracts have been provided (without annotations).

\paragraph{Train/dev split}
Given that no development set had been explicitly provided, we divided the PubMed ids of the original training set into a smaller training and a development set using an 80/20 split, in order to be able to prevent overfitting and perform early stopping. We assigned Pubmed ids to each one of the two sets in the prespecified proportions randomly but choosing a random seed that ensures a small KL divergence between the train and dev class distributions. In the rest of the paper, we use the terms `train set' or `training data' to refer to 80\% of the original annotated data that we use to train our model.

\paragraph{Generation of negative labels}

The training data contain some Pubmed ids that have no relation annotations whatsoever (either from the four pre-defined classes or explicitly negative). However, negative examples are crucial for training a model for such a task given that the majority of gene-disease pair mentions that are found in a randomly selected abstract are not expected to be related with a function change relationship. To generate pairs of negative mentions, we used a widely available Named Entity Recognition (NER) and Entity Linking system (see Section \ref{section:method}) to find mentions of genes and diseases in the abstracts. An entity mention predicted by NER was aligned to a labelled entity mention in the training data if they are both grounded to the same identifier. A pair was aligned if both its entity mentions (gene and disease) could be aligned. In less than 20\% of the pairs we performed manual alignment in order to improve the training signal. The dev set, however, was kept intact to ensure strict evaluation. The resulting distribution of relations is highly skewed towards the negative labels (`No relation'); the training set has the following distribution \texttt{(No relation: 0.939, GOF: 0.017, LOF: 0.03, REG: 0.012, COM: 0.0007)} while for the dev set, it is \texttt{(No relation: 0.935, LOF: 0.027, GOF: 0.019, REG: 0.016, COM: 0.003)}. `COM` is the least represented relationship with only two examples in the train set and two in the dev set. 

\section{Method}\label{section:method} 
This task can be decomposed into an NER step to obtain all gene-disease mention pairs in an abstract followed by a relation extraction (RE) step to predict the relation type for each mention pair found. 

For NER, we use Pubtator \cite{pubtator} to recognise spans tagged as genes or diseases. The main focus of our paper is performing relation extraction given NER labels. The reported results, however, don't assume gold NER labels.

\paragraph{Relation Extraction Model}
Our model is a simple extension of SciBERT \cite{scibert} for use in relation extraction, inspired by the encoding of mention pairs and textual context used in \cite{tre-transformer-re}. SciBERT, which utilises the same model architecture as BERT-base, consists of 12 stacked transformer encoders each with 12 attention heads. It is pre-trained using two objectives: Masked language modelling (Cloze task \cite{cloze-procedure}) and next sentence prediction. When trained, it is provided with sentence pairs represented as follows: \texttt{[CLS] This is sentence 1 \texttt[SEP] This is sentence 2 [SEP]}. The \texttt{[SEP]} token indicates when each sequence ends. The final hidden state of the \texttt{[CLS]} token is fed into a classifier to predict whether the two sentences appear sequentially in the corpus. As a result, the final hidden state of the \texttt{[CLS]} token learns a representation of the relationship between the two input sentences. 

We adapt SciBERT for relation extraction by fine-tuning the representation learnt by the \texttt{[CLS]} hidden state. We encode each pair of gene-disease mentions along with the corresponding PubMed abstract in the following format: \texttt{[CLS] gene-mention disease-mention [SEP] This is the PubMed abstract [SEP]}. This input data is fed into SciBERT and the final hidden state of its \texttt{[CLS]} token is passed into a single linear layer to predict the relation type expressed in that abstract for that gene-disease mention pair. The \texttt{[CLS]} hidden state which was pre-trained to learn a representation of the relationship between two sentences is now fine-tuned to learn which relationship class exists between a gene-disease pair (first `sentence') and a PubMed abstract (second `sentence'). Our encoding is similar to the approach proposed in \cite{tre-transformer-re}. This adaptation, while simple, is powerful because it is completely agnostic to domain-specific idiosyncrasies; for example, it can be used for any entity types and relation labels. Further, as it has already been pre-trained on a large unstructured corpus, it can be fine-tuned using a considerably smaller dataset. 

\paragraph{Model training}
We use negative log likelihood of the true labels as a loss function. We train for at most 40 epochs with early stopping based on the dev set performance. We used two early stopping criteria alternatives: the macro-averaged F1-score over all labels and over just the positive labels. Training stops if the score used as stopping criterion does not increase on the dev set for 10 consecutive epochs or the maximum number of epochs has been reached. The batch size is chosen to be 32 and the maximum sequence length of each input sequence is set to be 350 Wordpiece (subword) tokens. This is due to memory constraints. For each batch, we used down-sampling to ensure that each class was represented equally on average. When training, we observed that our results were very sensitive to the classifier layer weight initialisations. This same behaviour was reported in the original BERT paper \cite{bert}. To address this, we performed 20 random restarts and selected the model that performs the best on the dev set (for each of the two stopping criteria).

\section{Experiments and results}
We report the standard classification metrics on the dev set: precision ($\bf P$), recall ($\bf R$), and F1-score ($\bf F1$). For each one of these metrics, we include the macro-averaged values, the micro-averaged values \textbf{i)} over all relation labels and \textbf{ii)} restricted to just the positive ones. We also report the per-class values (in a one-vs-all fashion). The best results are shown for both of the early stopping criteria used (see Tables \ref{table:model-results-positive} and \ref{table:model-results-all}).

\begin{table}[t!]
\begin{center}
\begin{tabular}{|l|llll|}
\hline \bf  & \bf P & \bf R & \bf F1 & \bf Supp.\\ \hline
No rel & 0.934  & 0.372 & 0.532 & 627 \\
REG & 0.174  & 0.087 & 0.116 & 11 \\
COM & 0  & 0 & 0 & 2 \\
LOF & 0.076  & 0.307 & 0.122 & 19\\
GOF & 0.022  & 0.577 & 0.042 & 12 \\ \hline
Micro-all & 0.368 &  0.368 & 0.368 &\\
Macro-all & 0.241 & 0.268 & 0.162 &\\
Micro-pos & 0.033 &  0.322 & 0.060 &\\
Macro-pos & 0.068 & 0.243 & 0.070 &\\
\hline
\end{tabular}
\end{center}
\caption{Model results on the four pre-defined classes, as well as `No rel' (the negative class) when the macro-averaged F1-score (over the positive labels only) is used as our early stopping criterion. P, R and F1 stand for Precision, Recall and F1-score respectively; $\mbox{support} = \mbox{true positives} + \mbox{false negatives}$. Micro-all and Macro-all are the micro- and macro-averaged metrics for all classes while Micro-pos and Macro-pos are the micro- and macro-averaged metrics for only the positive classes (i.e. four classes excluding `No rel').}
\label{table:model-results-positive}
\end{table}

\begin{table}[t!] 
\begin{center}
\begin{tabular}{|l|llll|}
\hline \bf  & \bf P & \bf R & \bf F1 & \bf Supp.\\ \hline
No rel & 0.937  & 0.761 & 0.840 & 627 \\
REG & 0  & 0 & 0 & 11 \\
COM & 0  & 0 & 0 & 2 \\
LOF & 0.214  & 0.040 & 0.067 & 19\\
GOF & 0.038  & 0.429 & 0.070 & 12 \\ \hline
Micro-all & 0.722 &  0.722 & 0.722 &\\
Macro-all & 0.238 & 0.246 & 0.196 &\\
Micro-pos & 0.037 &  0.141 & 0.059 &\\
Macro-pos & 0.063 & 0.117 & 0.034 &\\
\hline
\end{tabular}
\end{center}
\caption{Model results on the four pre-defined classes, as well as `No rel' (the negative class) when the macro-averaged F1-score (over all labels) is used as our early stopping criterion. All terms used here as defined in Table \ref{table:model-results-positive}.} \label{table:model-results-all}
\end{table} 


\begin{table}[t!]
\begin{center}
\begin{tabular}{|l|llll|}
\hline \bf  & \bf P & \bf R & \bf F1 & \bf Supp.\\ \hline
No rel & 0.934  & 0.92 & 0.927 & 627 \\
REG & 0  & 0 & 0 & 11 \\
COM & 0  & 0 & 0 & 2 \\
LOF & 0.043  & 0.053 & 0.048 & 19\\
GOF & 0  & 0 & 0 & 12 \\ \hline
Micro-all & 0.862 &  0.863 & 0.863 &\\
Macro-all & 0.195 & 0.195 & 0.195 &\\
Micro-pos & 0.019 &  0.023 & 0.021 &\\
Macro-pos & 0.011 & 0.013 & 0.012 &\\
\hline
\end{tabular}
\end{center}
\caption{Baseline results on the four pre-defined classes, as well as `No rel' (the negative class). All terms used here as defined in Table \ref{table:model-results-positive}.}
\label{table:baseline-results}
\end{table}

\paragraph{Random sampling-based baseline} 
We compare our model performance against a simple baseline that predicts the class label by sampling from the categorical distribution of labels as calculated from the training set. Given the strongly skewed class distribution (which has low entropy of 0.46 bits, compared to 2.32 bits for a 5-class uniform distribution, and is therefore highly predictable), this is a strong baseline, especially for metrics reported on frequent classes. Table \ref{table:baseline-results} summarises the results, which have been averaged over 1,000 random sampling experiments. As expected, all metrics can achieve high scores on the negative (and by far the largest) class, illustrating how misleading micro-averaging with large classes can be as an indicator of model performance. Some classes have zero scores, which is unsurprising given their very low support in the dev set.

\paragraph{Discussion} 
For both early stopping criteria mentioned above, our model significantly outperformed the random baseline on macro-averaged metrics and per-class metrics. The model obtained relatively good performance on the positive labels especially when taking into account the considerable class imbalance. When optimised to the macro-averaged F1-score over just the positive labels, the model performance was unsurprisingly slightly superior over the positive labels compared to when optimised using the macro-averaged F1-score over all labels. However, this came at the expense of a loss in recall on the negative labels. To generate predictions on the test set, we chose the model optimised using the macro-averaged F1-score over just the positive labels.

\paragraph{Pubtator NER performance}
The performance of our relation extraction model is dependent on the results of the named entity recognition tool. Here we briefly summarise the performance of the Pubtator NER tool on the dev set. There are 44 entity pairs with positive labels in the dev set. Of these 44, Pubtator correctly identified 24 of them with an exact string match. For the remaining 20, 14 were identified but it was not an exact string match, and for the other 6, at least one of the entities was not found. We were fairly strict for our dev set evaluation, and so unless there was a perfect string match, the entities were not considered aligned to the labelled data. This would have degraded our performance metrics.

\section{Related work}

Many biomedical relation extraction systems have often relied hand-crafted linguistic features \cite{cdr-linguistic-features,cdr-rich-features} but recently also convolutional neural networks \cite{cnn-cdr,cnn-ppi}, LSTM \cite{neural-joint-bio-re,lstm-ddi} or a combination of machine learning models and neural-network-based encoders \cite{hybrid-neural-bio-re,ensemble-svm-cnn-rnn}. A recent paper \cite{bran} achieves state-of-the-art results on biomedical relation classification for chemically-induced diseases (CDR \cite{cdr}) and ChemProt (CPR \cite{chemprot}), by using a Transformer encoder \cite{attention-is-all-you-need} and end-to-end Named Entity Recognition and relation extraction, without, however, leveraging transformer-based language model pre-training. In the general domain, \cite{re-survey}
and \cite{re-distant-superv-survey} provide a comprehensive review of different relation extraction paradigms and methods that have been developed to date.

\section{Conclusions and further work}

We have presented a system that extracts mentions of biomedical entities and classifies them into one of four function change relations (or absence of a relation). Our system leverages widely available language representations pre-trained on biomedical data and utilises minimal task-specific architecture, while not relying on specially engineered linguistic features. Despite the model simplicity and the class imbalance in the data (even within the four non-negative classes), our model is able to significantly outperform the random baseline. 

Our model can be improved by using more recent language modeling methods, such as XLNet \cite{xlnet}, and different ways of encoding the mention pairs and textual context (e.g. by using not only the hidden state of the \texttt{[CLS]} token but also the hidden states of the entity mentions as input to the relationship classifier). Different methods can be explored for addressing class imbalance (e.g. a cost-sensitive classifier, data augmentation etc). Further, an end-to-end Named Entity Recognition and Relation Extraction architecture can be devised. It would also be interesting to compare our model against more competitive baselines.

\section*{Acknowledgments}
We would like to thank Nathan Patel for his engineering support as well as Angus Brayne, Julien Fauqueur and other colleagues working on NLP for insightful discussions.

\bibliographystyle{acl_natbib}

\end{document}